\documentclass[runningheads]{llncs}
\usepackage[T1]{fontenc}
\usepackage{hhline}
\usepackage{graphicx}
\usepackage{hyperref}
\usepackage[dvipsnames]{xcolor}
\usepackage[perpage]{footmisc}
\usepackage{amsmath}
\usepackage{array}
\usepackage{amssymb}
\usepackage[misc,geometry]{ifsym} 

\makeatletter
\def\thickhline{%
  \noalign{\ifnum0=`}\fi\hrule \@height \thickarrayrulewidth \futurelet
   \reserved@a\@xthickhline}
\def\@xthickhline{\ifx\reserved@a\thickhline
               \vskip\doublerulesep
               \vskip-\thickarrayrulewidth
             \fi
      \ifnum0=`{\fi}}
\makeatother

\makeatletter
\newlength{\thickarrayrulewidth}
\def\middlehline{%
  \noalign{\ifnum0=`}\fi\hrule \@height \middlearrayrulewidth \futurelet
   \reserved@a\@xmiddlehline}
\def\@xmiddlehline{\ifx\reserved@a\middlehline
               \vskip\doublerulesep
               \vskip-\middlearrayrulewidth
             \fi
      \ifnum0=`{\fi}}
\makeatother

\newlength{\middlearrayrulewidth}
\makeatletter
\def\blfootnote{\xdef\@thefnmark{}\@footnotetext}
\makeatother

\usepackage{color}

\begin{document}

\newcommand{\FA}{\operatorname{FA}}
\newcommand{\AD}{\operatorname{AD}}
\newcommand{\MD}{\operatorname{MD}}
\newcommand{\RD}{\operatorname{RD}}
\title{Enhancing Angular Resolution via Directionality Encoding and Geometric Constraints in Brain Diffusion Tensor Imaging}
\titlerunning{DirGeo-DTI for DTI Angular Enhancement}

\author{Sheng Chen\inst{1,2} \and
Zihao Tang\textsuperscript{1,2,3(\Letter)} \and
Mariano Cabezas\inst{2,3,4} \and
Xinyi Wang\inst{1,2} \and
Arkiev D'Souza \inst{2} \and
Michael Barnett\inst{2,4,6} \and
Fernando Calamante \inst{2,5,6} \and
Weidong Cai\inst{1} \and
Chenyu Wang\inst{2,3,6}}

\institute{School of Computer Science, University of Sydney, NSW 2008, Australia \\
\email{zihao.tang@sydney.edu.au}\and
Brain and Mind Centre, University of Sydney, NSW 2050, Australia\and
Central Clinical School, University of Sydney, NSW 2050, Australia\and
Sydney Neuroimaging Analysis Centre, University of Sydney, NSW 2050, Australia\and
School of Biomedical Engineering, University of Sydney, NSW 2008, Australia\and
Sydney Imaging, University of Sydney, NSW 2006, Australia\\}

\authorrunning{S. Chen et al.}
\maketitle              

\begin{abstract}
Diffusion-weighted imaging (DWI) is a type of Magnetic Resonance Imaging (MRI) technique sensitised to the diffusivity of water molecules, offering the capability to inspect tissue microstructures and is the only in-vivo method to reconstruct white matter fiber tracts non-invasively. The DWI signal can be analysed with the diffusion tensor imaging (DTI) model to estimate the directionality of water diffusion within voxels. Several scalar metrics, including axial diffusivity (AD), mean diffusivity (MD), radial diffusivity (RD), and fractional anisotropy (FA), can be further derived from DTI to quantitatively summarise the microstructural integrity of brain tissue. These scalar metrics have played an important role in understanding the organisation and health of brain tissue at a microscopic level in clinical studies. However, reliable DTI metrics rely on DWI acquisitions with high gradient directions, which often go beyond the commonly used clinical protocols. To enhance the utility of clinically acquired DWI and save scanning time for robust DTI analysis, this work proposes DirGeo-DTI, a deep learning-based method to estimate reliable DTI metrics even from a set of DWIs acquired with the minimum theoretical number (6) of gradient directions. DirGeo-DTI leverages directional encoding and geometric constraints to facilitate the training process. Two public DWI datasets were used for evaluation, demonstrating the effectiveness of the proposed method. Extensive experimental results show that the proposed method achieves the best performance compared to existing DTI enhancement methods and potentially reveals further clinical insights with routine clinical DWI scans. The code of the proposed DirGeo-DTI is available at \url{https://mri-synthesis.github.io/}.

\end{abstract}

\section{Introduction}
Magnetic Resonance Imaging (MRI) is a non-invasive imaging technique based on the principles of nuclear magnetic resonance, used to generate detailed internal images of the human body~\cite{mcrobbie2017mri,tang2023high}. Among MRI techniques, diffusion-weighted imaging (DWI) is sensitised to Brownian motion and measures water diffusivity as it is constrained by various tissue structures~\cite{tournier2011diffusion}. Thus, DWI can reveal microstructural organisation and assess the impact of pathology on tissue integrity. Therefore, it has become a remarkable surrogate in clinical neuroscience research since the 1990s~\cite{tsuruda1990diffusion,assaf2019role}. 

The Diffusion Tensor Imaging (DTI), one of the most widely used DWI models, characterised by a tensor matrix $\textbf{D}$ as follows:
\begin{align}
\label{eq:dti}
\textbf{D} = 
\begin{bmatrix}
D_{xx} & D_{xy} & D_{xz} \\
D_{xy} & D_{yy} & D_{yz} \\
D_{xz} & D_{yz} & D_{zz} \\
\end{bmatrix}.
\end{align}
The symmetry of $\textbf{D}$ implies that, theoretically, only 6 unique coefficients need to be estimated, requiring at least 6 unique DWIs. In clinical research studies, a greater number of DWIs are generally used to ensure an accurate DTI estimation~\cite{ni2006effects}. Specifically, it was recommended to acquire more than 20 diffusion gradient directions to obtain reliable DTI scalar metrics~\cite{jones2004effect}. However, DWI has been much more commonly used for stroke diagnosis in clinical practice~\cite{gaddamanugu2022clinical} and has broad applications to differentiate pathologies that result in diffusion restriction, which is often acquired with much fewer DWI directions. While leveraging retrospective clinical data for new observational studies, DWIs obtained through clinical protocols might not include sufficient gradient directions for reliable DTI metrics. To maximize the utility of clinically acquired DWI in the aforementioned scenarios, deep learning approaches~\cite{tang2023high,tian2020deepdti} have been developed to enhance the angular resolution of DTI, aiming to derive reliable DTI metrics from a limited number of DWIs. However, these methods have yet to integrate the $b_{vec}$, a critical imaging parameter that contains valuable directional information for angular resolution enhancement. Furthermore, they have ignored exploiting the inherent geometrical properties of DTI model, which can be leveraged to constrain the differences between the enhanced DTI and the ground truth. To this end, we propose a DirGeo-DTI, which enhances the angular resolution of DTI with only 6 unique diffusion gradient directions via directionality encoding and geometric constraints. Experimental results on two public datasets demonstrate that DirGeo-DTI outperforms existing methods with scalar metrics derived from the enhanced DTIs by DirGeo-DTI comparable to those derived from the ground truth. Our contributions can be summarized as follows:
\begin{itemize}
    \item DirGeo-DTI is the first method to integrate the information of diffusion gradient direction ($b_{vec}$) by the proposed novel Diffusion Gradient Encoding (DGE). 
    \item DirGeo-DTI has leveraged geometric learning, including stress invariants from physics and FA, to constrain the geometrical properties of the predicted DTI model.
    \item Extensive experimental results demonstrate that DirGeo-DTI achieves the best performance and has the potential to reveal further clinical insights with routine clinical DWI scans.
\end{itemize}

\section{Methods}
\subsection{Related Works}
In recent years, deep learning has significantly advanced structural brain MRI applications, leading to successful translation into real clinical practice~\cite{barnett2023ai,tang2021lg}. DWI, in contrast, the complexity of data requires sophisticated techniques for accurate analysis and reconstruction. Several deep learning-based methods have been proposed to address various challenges in this domain. To maximize the utility of retrospective DWI, efforts have been made to reconstruct missing slices caused by the limited  field of view (FOV) in acquisition protocols through deep learning techniques~\cite{tang75reducing,tang2022tw,gao2024field}. Clinically acquired scans are often constrained by acquisition protocols with low spatial and angular resolutions due to cost limitations~\cite{aja2023validation}. Several methods were proposed to perform spatial super-resolution on DWI and its derivatives. While angular resolution is a property unique to diffusion MRI scans, few studies have specifically addressed this aspect. DeepDTI~\cite{tian2020deepdti} and HADTI-Net~\cite{tang2023high} were developed to enhance the angular resolution of DTI. DeepDTI utilises raw DWI, b0, and T1/T2-weighted images as inputs, which are processed by a 3D convolutional neural network (CNN) to improve DWIs, followed by fitting the enhanced images to a tensor model. HADTI-Net, on the other hand, employs a 3D U-Net to use DWI, b0, and T1-weighted images to directly produce enhanced DTI metrics. While both methods have demonstrated improved DTI quality, they have not fully investigated the impact of incorporating diffusion gradient directions and geometric constraints in enhancing the angular resolution of DTI.

\subsection{Datasets}
The Human Connectome Project (HCP) and Parkinson's Progression Markers Initiative (PPMI) databases were used for this study.

The HCP database~\cite{van2013wu} comprises structural imaging and DWI data acquired with 3 diffusion-weighted shells ($b_{1000}, b_{2000}$ and $b_{3000}$ with 90 gradient directions for each b-value). We randomly selected 100 subjects from the HCP database and split them between training (80) and testing (20) in this work. 

\begin{figure*}[!htb]
\centering
\includegraphics[width=\linewidth]{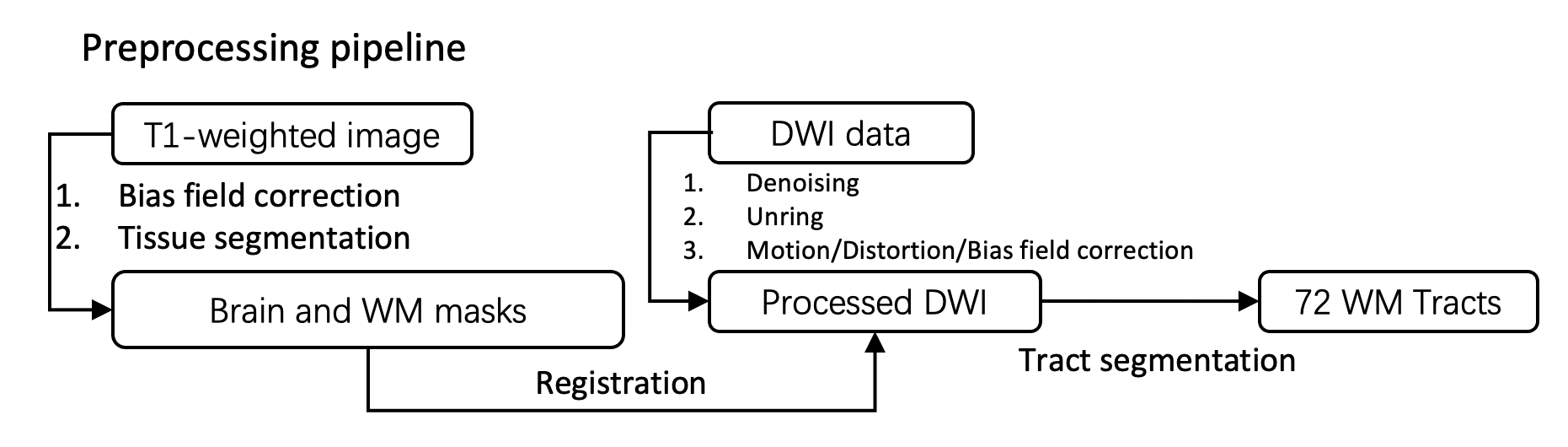}
\caption{The preprocessing pipeline of raw diffusion and T1 images. Image preprocessing including denoising, unring, and corrections of bias field, motion, and distortion were performed using MRtrix3~\cite{tournier2019mrtrix3}. Tissue and WM tract segmentation were conducted with~\cite{fischl2012freesurfer} and ~\cite{wasserthal2018tractseg}, respectively.}
\label{fig:prep}
\end{figure*}

PPMI~\cite{marek2011parkinson} is a collaborative study for Parkinson's Disease (PD) that comprises a large-scale imaging data collection including DWIs and structural imaging for both healthy controls and patients with PD. For each subject, DWI scans with 64 gradient directions were acquired with $b_{1000}$. In our study, 225 subjects from the dataset were selected for model training (180) and testing (45). The class balance was maintained for both the training and testing sets. In contrast with HCP whose data is already pre-processed, the images required additional preprocessing for the raw DWI~\cite{andersson2015non,andersson2016integrated,wasserthal2018tractseg,tournier2019mrtrix3} and T1~\cite{jenkinson2012fsl,fischl2012freesurfer} images as illustrated in Figure~\ref{fig:prep}.

\subsection{Data Preparation}
All available $b_{1000}$ directions from the preprocessed DWIs in each dataset were first selected to derive the ground truth DTI due to the signal diminishing at high b-values~\cite {jensen2010mri} and $b_{1000}$ is commonly used in clinical acquisitions. From this set of images, 6 most evenly distributed directions from all possible sets were sub-sampled using the Kennard-Stone algorithm~\cite{kennard1969computer} in Q-Space to select the most evenly distributed set of 6-direction DWIs as input to the proposed network. A diffusion tensor model was then fitted at each voxel~\cite{behrens2003characterization} of the two sets of preprocessed images using FSL DTIFIT to generate the corresponding DTI volumes for the 6 direction volumes (namely 6-dir DTI) and the ground truth ones $\mathbf{D_{GT}}$ (derived from all available $b_{1000}$ directions). 72 White Matter (WM) fiber tracts were generated using TractSeg~\cite{wasserthal2018tractseg} on the original preprocessed DWIs.

\subsection{DirGeo-DTI}
\begin{figure*}[!htb]
\centering
\includegraphics[width=\linewidth]{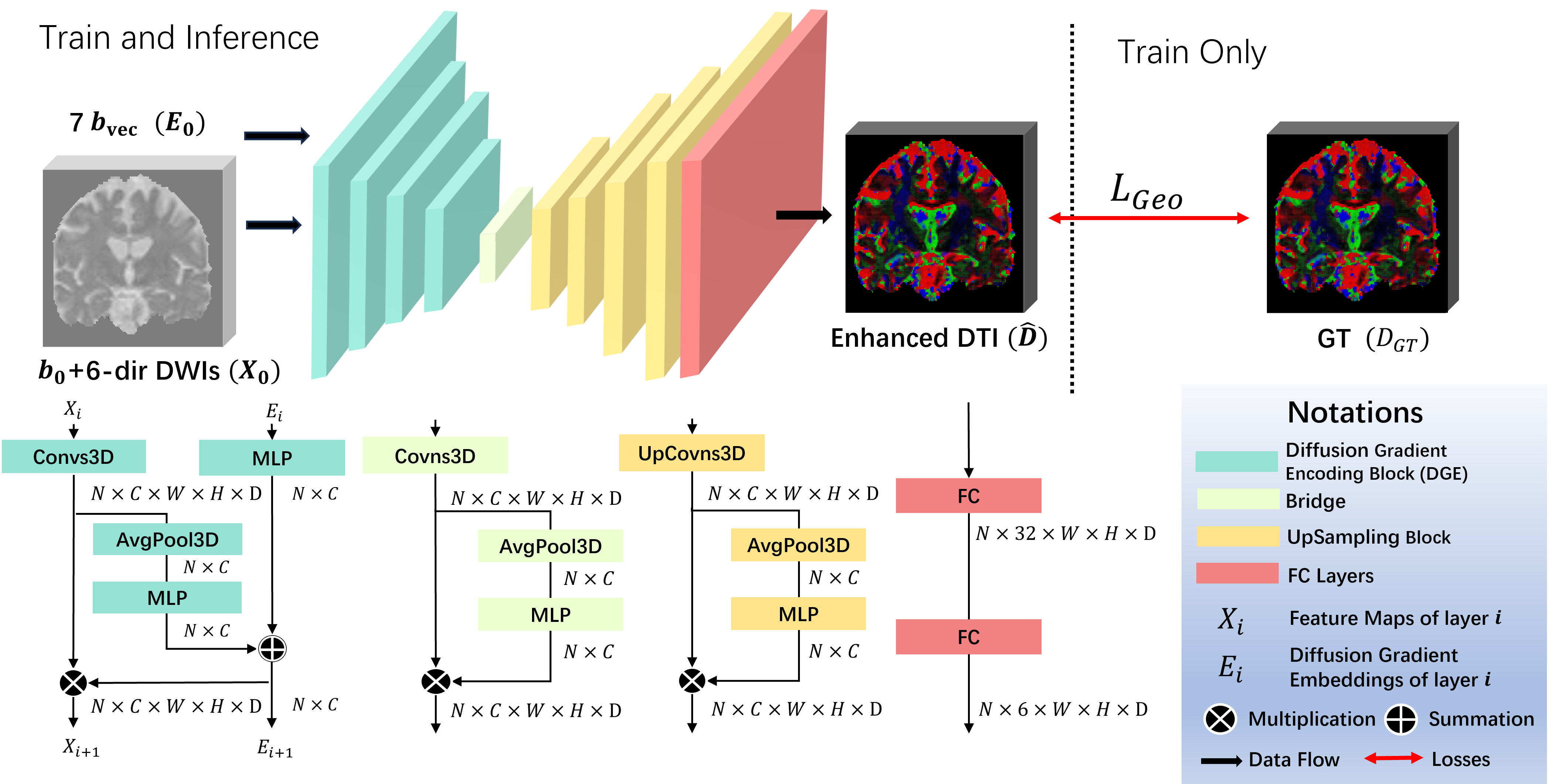}
   \caption{Detailed framework of the proposed DirGeo-DTI.}
\label{fig:net}
\end{figure*}

The detailed design of the proposed DirGeo-DTI is illustrated in Figure~\ref{fig:net}. DirGeo-DTI takes $b_0$ image plus 6 unique DWIs and 7 corresponding $b_{vec}$ ($(0,0,0)$ as $b_{vec}$ for $b_0$ image) as input to generate enhanced DTI volumes ($\mathbf{\hat{D}}$). 

\subsubsection{Directionality Encoding}
The diffusion tensor $\textbf{D}$ at each voxel can be estimated from DWI data using the Stejskal-Tanner equation: $S_{k}=S_{0}e^{-b \hat{g}^T_k \textbf{D} \hat{g}_{k}}$~\cite{stejskal1965spin}. In this equation, $S_{0}$ represents the signal acquired without any diffusion-sensitizing gradients (often referred to as the $b_{0}$ image), while $S_{k}$ represents the diffusion-weighted signal measured along the gradient direction $\hat{g}_{k}$ with diffusion weighting factor $b$~\cite{basser1995inferring,o2011introduction}. The gradient direction $\hat{g}_{k}$, commonly referred to as $b_{vec}$, is a crucial parameter in diffusion MRI acquisition and DTI modeling, providing essential directional information to solve the Stejskal-Tanner equation. For a given gradient direction, $\hat{g}_{k}= (g_x, g_y, g_z)$, the Stejskal-Tanner equation can be expanded to:
\begin{align}
\label{eq:st}
\frac{\ln(S_0) - \ln(S_k)}{b} = \mathbf{g} \cdot \mathbf{d},
\end{align}
where $\mathbf{g}$ is defined as the vector $[g^{2}_{x}, 2g_{x}g_{y}, 2g_{x}g_{z}, g^{2}_{y}, 2g_{y}g_{z}, 
g^{2}_{z}]$ and $\mathbf{d}$ represents our estimation target $[D_{xx}, D_{xy}, D_{xz}, D_{yy}, D_{yz}, D_{zz}]$. 

Inspecting the Equation~\ref{eq:st}, leveraging $b_{vec}$ is vital for enhancing the angular resolution of DTI because it encodes directional constraints. A novel Diffusion Gradient Encoding (DGE) block is thus proposed to leverage $b_{vec}$ efficiently. The DGE block is designed to incorporate directional information directly into the network's learning process. It takes the feature map $X_i \in \mathbb{R}^{N\times C\times W \times H \times D}$ and a diffusion gradient embedding $E_{i} \in \mathbb{R}^{N\times C}$ as inputs, where $i$ denotes the layer position within the network. The DGE block processes these inputs through a multilayer perceptron (MLP) that maps the diffusion-weighted information from $X_i$ and $E_{i}$ into latent embeddings separately. These two embeddings are then combined and used to recalibrate the feature map channels in the subsequent layers~\cite{hu2018squeeze}. The overall process in the DGE block 
can be formulated as follows:
\begin{equation}
E_{i+1} = \mathcal{F}_{2}(\mu(\mathcal{F}_{1}(X_i))) + \mathcal{F}_{3}(E_{i}),
\end{equation}
\begin{equation}
X_{i+1} = \mathcal{F}_{1}(X_i) \times E_{i+1},
\end{equation}
where $\mathcal{F}_{1}$ is a convolution operation, $\mathcal{F}_{2}$ and $\mathcal{F}_{3}$ are two independent MLPs, and $\mu$ denotes average pooling. This design ensures that the directional information from the gradient directions is effectively integrated into the feature extraction process, providing a more robust representation of diffusion information and improving the estimation of the corresponding DTI model. The DGE block, therefore, learns a representation of the missing diffusion gradient directions and signals in a latent space, enhancing the angular resolution of the estimated DTI. This novel approach of incorporating $b_{vec}$ directly into the network's architecture is a key innovation of our proposed DirGeo-DTI framework.

\subsection{Geometric Learning}
\subsubsection{Stress Invariants}
From a physics perspective, the mathematical representation of $\textbf{D}$ describes a stress tensor ($\sigma$) representing directional variations in stress due to stress anisotropies, as shown in Equation~\ref{eq:stress}. Moreover, stress invariants associated with $\sigma$ are a set of constraints that are unaffected by stress and are important parameters used in failure criteria~\cite{camanho2015three,hashin1980fatigue}, plasticity, and Mohr's circle~\cite{krenk1996family}. In DTI angular resolution enhancement, stress invariants can be used to constrain the enhanced DTI prediction and force geometrical similarity with the corresponding ground truth. Based on this observation, we integrate the stress invariants of $\sigma$ as part of the geometric learning of DirGeo-DTI. 

Specifically, the stress invariants of a symmetric $3\times3$ matrix can be expressed by the three different orders of principle minors of the tensor. The first-order principal minors ($\Delta_{1}$) are defined by the diagonal elements of the stress tensor ($\sigma_{xx}$, $\sigma_{yy}$, and $\sigma_{zz}$) and represent the normal stresses on the three axes. This constraint in $\Delta_{1}$, can be easily covered by calculating the L1 loss between GT ($\mathbf{D_{GT}}$) and the enhanced DTI ($\mathbf{\hat{D}}$) from the network:
\begin{equation}
    \mathcal{L}_{DTI} = ||\mathbf{D_{GT}} - \mathbf{\hat{D}}||_{1}.
\end{equation}
The second-order principal minors ($\Delta_{2}$) represent the contributions of the shear stress components to the overall stress state, reflecting the interaction between normal and shear stresses on different planes. The definitions for $\Delta_{2}$ are formulated in Equations~\ref{eq:pm2_1} and~\ref{eq:pm2_2} and a new loss $\mathcal{L}_{\Delta_{2}}$ is then implemented as Equation~\ref{eq:lpm2}.
\begin{equation}
\label{eq:stress}
\sigma = 
\begin{bmatrix}
\sigma_{xx} & \sigma_{xy} & \sigma_{xz} \\
\sigma_{xy} & \sigma_{yy} & \sigma_{yz} \\
\sigma_{xz} & \sigma_{yz} & \sigma_{zz}
\end{bmatrix},
\end{equation}
\begin{align}
\label{eq:pm2_1}
M_{1} &= 
\begin{bmatrix}
\sigma_{xx} & \sigma_{xy}\\
\sigma_{yx} & \sigma_{yy}
\end{bmatrix}&
M_{2} &= 
\begin{bmatrix}
\sigma_{xx} & \sigma_{xz}\\
\sigma_{zx} & \sigma_{zz}
\end{bmatrix} &
M_{3} &= 
\begin{bmatrix}
\sigma_{yy} & \sigma_{yz}\\
\sigma_{zy} & \sigma_{zz}
\end{bmatrix},
\end{align}
\begin{equation}
\label{eq:pm2_2}
    \Delta_{2}(\sigma) = det(M_{1}) + det(M_{2}) + det(M_{3}),
\end{equation}
\begin{equation}
\label{eq:lpm2}
    \mathcal{L}_{\Delta_{2}} = ||\Delta_{2}(\mathbf{D_{GT}}) - \Delta_{2}(\mathbf{\hat{D}})||_{1}.
\end{equation}
Finally, the third-order principal minors ($\Delta_{3}$) are defined as the determinant of the entire stress tensor. From a theoretical perspective, DTI at a certain voxel in the brain can be understood as applying a 3D linear transformation (defined by rotation, scaling, reﬂection, and shear) to a sphere with its origin point at its centre. In addition, DTI is generally visualised as a 3D ellipsoid defined by $\textbf{D}$, where the lengths of the semi-principle axes of the ellipsoid are defined by the eigenvalues of $\textbf{D}$. We thus leverage the definition of $\Delta_{3}$ to constrain the volume invariant between the ellipsoids represented by $\mathbf{D_{GT}}$ and $\mathbf{\hat{D}}$ by calculating the relative volume change:
\begin{equation}
    \rho_{v} = \frac{\Delta_{3}(\mathbf{D_{GT}})}{\Delta_{3}(\mathbf{\hat{D}})} = \frac{det(\mathbf{D_{GT}})}{det(\mathbf{\hat{D}})} .
\end{equation}
We intend to consistently penalise the percentage volume differences by multiplying the original loss by a penalisation factor $\xi$, which is directly proportional to the percentage change in volume and follows $\xi \geq 1$:
\begin{equation}
\label{eq:lambdav}
    \xi = \begin{cases} 
        \rho_{v} & \text{if } \rho_{v} \geq 1, \\
        \frac{1}{\rho_{v}} & \text{if } \rho_{v} < 1,
        \end{cases}
\end{equation}
$\xi$ is thus implemented as a single penalisation factor for back-propagation:
\begin{equation}
    \xi = \frac{1}{\rho_{v}} ReLU (1-\rho_{v}) + ReLU(\rho_{v}-1) + 1. 
\end{equation}

\subsubsection{Relative Diffusivity}
To provide quantitative insights into the microstructural characteristics of the underlying brain tissue~\cite{o2011introduction}, scalar metrics, including axial diffusivity (AD), mean diffusivity (MD), radial diffusivity (RD), and fractional anisotropy (FA) can then be derived from $\textbf{D}$. As demonstrated in Equation \ref{eq:admdrd} and \ref{eq:fa}, these metrics are similar to $\Delta_{3}$ that is closely related to the eigenvalues $(\lambda_{1}, \lambda_{2}, \lambda_{3})$ of $\textbf{D}$, measuring the magnitude and morphology of the ellipsoid represented by the tensor and serving as effective imaging biomarkers. 
\begin{equation}
\AD = \lambda_1, \MD = \frac{\lambda_1 + \lambda_2 + \lambda_3}{3}, \RD = \frac{\lambda_2 + \lambda_3}{2},
\label{eq:admdrd}
\end{equation}
\begin{equation}
\FA = \sqrt{\frac{(\lambda_1-\lambda_2)^2 + (\lambda_1-\lambda_3)^2 + (\lambda_2-\lambda_3)^2}{2*(\lambda_1^2 + \lambda_2^2 + \lambda_3^2)}}.
\label{eq:fa}
\end{equation}
$\FA$ is a ratio of variances normalised by the total diffusion that naturally falls within the scale of 0 to 1, making it a dimensionless measure in contrast to scale measures ($\AD$, $\MD$, and $\RD$). Therefore, based on the significance and properties of $\FA$, we adopt it as a complementary objective function for geometric learning:
\begin{align}
\label{eq:fa_loss}
   \mathcal{L}_{FA} = ||\FA({\mathbf{D_{GT}}}) - \FA({\mathbf{\hat{D}}})||_{1}.
\end{align}
\subsubsection{Geometry-constrained Loss}
Hence, the total geometry-constrained loss can be formulated as:
\begin{equation}
    \mathcal{L}_{Geo} = \xi (\alpha \mathcal{L}_{DTI} + \beta \mathcal{L}_{\Delta_{2}}) + \gamma \mathcal{L}_{FA},
\end{equation}
where $\alpha$, $\beta$, and $\gamma$ serve to balance the contributions of both loss items by adjusting the numerical magnitude of them to the range of $[0, 1]$.

\subsection{Implementation Details}
3D U-Net~\cite{cciccek20163d} is adopted as a baseline. All compared deep learning methods were trained for 100 epochs for each dataset using Adam optimiser with an initial learning rate of 0.001. The preprocessed 6-direction DWIs were cropped into patches of $64\times64\times64$ with an overlap of $32\times32\times32$ and were fed into the network in batches of 12. The $b_{0}$, $b_{vec}$ (unit vectors), and preprocessed DWIs were not normalised for training and testing. Our model is optimized based on the weighted loss $\mathcal{L}_{Geo}$ using the weight values $\alpha = 1e6$, $\beta = 1e6$ and $\gamma = 10$, which were empirically chosen to yield optimum results. All experiments were run on a single NVIDIA GeForce Tesla V100-SXM2 GPU with 32GB of memory and the models were implemented with PyTorch 2.0.0 and Python 3.10.12.

\section{Experimental Results}
\subsection{Tensor Evaluation}
To quantitatively assess the differences between DTIs generated using various methods and the ground truth, the mean absolute error (MAE) of the DTI coefficients and the corresponding diffusion scalar metrics in white matter (WM) regions of interest are presented in Table~\ref{table:mae}.

\begin{table}[!htb]
\caption{\label{table:mae}Summary of average and standard deviation values for the mean absolute error between enhanced DTI and ground truth of common DTI-derived scalar metrics. The units of measurement for AD, MD, and RD are expressed in $mm^2/s$.}
\centering
\begin{tabular}{l|l|l|l|l|l}
\hline\hline
\multicolumn{6}{c}{\textbf{MAE $\downarrow$}} \\ \hline \hline
 & \multicolumn{5}{c}{\textbf{HCP}} \\ \hline 
Method & DTI $(10^{-5})$ & $\FA $ & $\MD (10^{-5}) $ & $\AD (10^{-5}) $ & $\RD (10^{-5}) $ \\ \hline

6-dir DTI & $24.5766_{16.4509}$ & $0.2678_{0.0836}$ & $5.3319_{0.0002}$ & $38.9953_{29.2846}$ & $19.9455_{14.7553}$ \\  \hline
DeepDTI~\cite{tian2020deepdti}& $14.1192_{1.8006}$ & $0.1339_{0.0077}$ & $5.0022_{2.7591}$ & $16.5967_{5.1187}$ & $7.7222_{1.3883}$ \\  \hline
HADTI-Net~\cite{tang2023high} & $6.2000_{0.5714}$ & $0.0628_{0.0115}$ & $4.4172_{1.6251}$ & $8.1080_{0.7112}$ & $5.4671_{1.8807}$ \\ \hline
Baseline~\cite{cciccek20163d} & $6.1226_{0.6393}$ & $0.0621_{0.0132}$ & $4.3315_{1.5731}$ & $8.2590_{0.6843}$ & $5.5625_{2.1104}$ \\  \hline
+Directionality & $5.8423_{0.5340}$ & $0.0567_{0.0092}$ & $3.9942_{1.2218}$ & $7.7058_{0.6967}$ & $5.1957_{1.5600}$ \\   \hline
+Geometry & $5.9985_{0.1491}$ & $0.0595_{0.0119}$ & $4.6778_{1.4829}$ & $7.8186_{0.5904}$ & $5.8982_{1.9099}$\\  \hline
Proposed & $\textbf{5.4347}_{0.4698}$ & $\textbf{0.0511}_{0.0078}$ & $\textbf{3.6938}_{0.8190}$ & $\textbf{7.4327}_{0.7940}$ & $\textbf{4.6645}_{1.0387}$ \\ \hline 
& \multicolumn{5}{c}{\textbf{PPMI}} \\ \hline

6-dir DTI & $11.5758_{3.2577}$ & $0.1408_{0.0458}$ & $7.8892_{2.6067}$ & $15.7595_{5.6056}$ & $8.9901_{2.8411}$ \\  \hline
DeepDTI & $8.3027_{1.7142}$ & $0.0905_{0.0231}$ & $7.5812_{3.1247}$ & $10.5421_{3.2736}$ & $8.4566_{2.5731}$ \\  \hline
HADTI-Net & $7.1588_{1.0795}$ & $0.0702_{0.0144}$ & $7.4638_{2.6458}$ & $9.4707_{1.3571}$ & $8.3025_{2.7729}$ \\ \hline
Baseline & $7.0102_{1.4342}$ & $0.0661_{0.0146}$ & $7.2151_{3.2949}$ & $8.8892_{1.7894}$ &  $8.3604_{3.6755}$ \\ \hline
+Directionality  & $7.0005_{1.5468}$ & $0.0572_{0.0113}$ & $7.4415_{3.6142}$ & $9.3485_{2.3807}$ & $8.2072_{3.6495}$  \\ \hline
+Geometry & $6.8898_{1.4912}$ & $0.0636_{0.0153}$ & $7.3639_{3.4444}$ & $8.8921_{1.8161}$ & $8.5012_{3.8544}$ \\ \hline
Proposed & $\textbf{6.2649}_{1.0594}$& $\textbf{0.0534}_{0.0090}$& $\textbf{6.0502}_{2.3102}$& $\textbf{8.3576}_{1.8976}$& $\textbf{6.6726}_{2.1910}$ \\ \hline
\end{tabular}
\end{table}

\begin{table}[!htb]
\caption{\label{SSIM}Summary of average and standard deviation values for the Structural Similarity Index (SSIM) between enhanced DTI and ground truth of common DTI-derived scalar metrics.}
\centering
\begin{tabular}{l|l|l|l|l|l}
\hline \hline
\multicolumn{6}{c}{\textbf{SSIM $\uparrow$}} \\ \hline \hline
& \multicolumn{5}{c}{\textbf{HCP}} \\ \hline 
Method & DTI  & FA  & MD  & AD  & RD  \\ \hline
6-dir DTI & $0.2581_{0.1119}$& $0.2207_{0.0766}$& $0.5990_{0.1229}$& $0.2770_{0.1049}$& $0.2935_{0.1113}$\\  \hline
DeepDTI& $0.2573_{1.8006}$& $0.5533_{0.0128}$& $0.7674_{0.0361}$& $0.5467_{0.0335}$& $0.6334_{0.0241}$\\  \hline
HADTI-Net & $0.6274_{0.0243}$& $0.7663_{0.0263}$& $0.7818_{0.0405}$& $0.7229_{0.0335}$& $0.7586_{0.0305}$\\ \hline
Baseline& $0.6468_{0.0273}$& $0.7733_{0.0295}$& $0.7815_{0.0407}$& $0.7296_{0.0346}$& $0.7603_{0.0214}$\\  \hline
+Directionality & $0.6314_{0.0227}$&  $0.7915_{0.0250}$& $0.7808_{0.0381}$& $0.7375_{0.0323}$& $0.7626_{0.0287}$\\   \hline
+Geometry & $0.6388_{0.1491}$& $0.7840_{0.0265}$& $0.7677_{0.0337}$& $0.7378_{0.0293}$& $0.7496_{1.9099}$\\  \hline
Proposed & $\textbf{0.6538}_{0.0249}$& $\textbf{0.8120}_{0.0233}$& $\textbf{0.7853}_{0.0464}$& $\textbf{0.7522}_{0.0309}$& $\textbf{0.7785}_{0.0328}$\\ \hline 
& \multicolumn{5}{c}{\textbf{PPMI}} \\ \hline

6-dir DTI & $0.4472_{0.0695}$& $0.5042_{0.0925}$& $0.9037_{0.0360}$& $0.5865_{0.1977}$& $0.7393_{0.0676}$\\  \hline
DeepDTI & $0.5091_{0.0489}$& $0.6549_{0.0324}$& $0.7067_{0.6064}$& $0.6425_{0.0306}$& $0.7550_{0.0345}$\\  \hline
HADTI-Net & $0.6041_{0.0256}$& $0.7591_{0.0374}$& $0.8821_{0.0214}$& $0.7837_{0.0179}$& $0.8272_{0.0625}$\\ \hline
Baseline & $0.6416_{0.0290}$& $0.7894_{0.0339}$& $0.8985_{0.0470}$& $0.7401_{0.4752}$&  $0.8697_{0.0350}$\\ \hline
+Directionality & $0.6524_{0.0281}$& $0.8120_{0.0297}$& $0.9055_{0.0358}$& $0.8148_{0.0265}$& $0.8775_{0.0349}$\\ \hline
+Geometry & $0.6642_{0.1139}$& $0.7910_{0.0369}$& $0.9051_{0.0369}$& $0.8138_{0.0269}$& $0.8709_{0.0353}$\\ \hline
Proposed & $\textbf{0.6651}_{0.0249}$& $\textbf{0.8226}_{0.0276}$& $\textbf{0.9081}_{0.0260}$& $\textbf{0.8229}_{0.0201}$& $\textbf{0.8869}_{0.8946}$\\ \hline
\end{tabular}
\end{table}

\begin{figure}[!htb]
\begin{minipage}[b]{0.18\linewidth}
  \centering
  \centerline{\includegraphics[width=2.4cm,height=3.0cm]{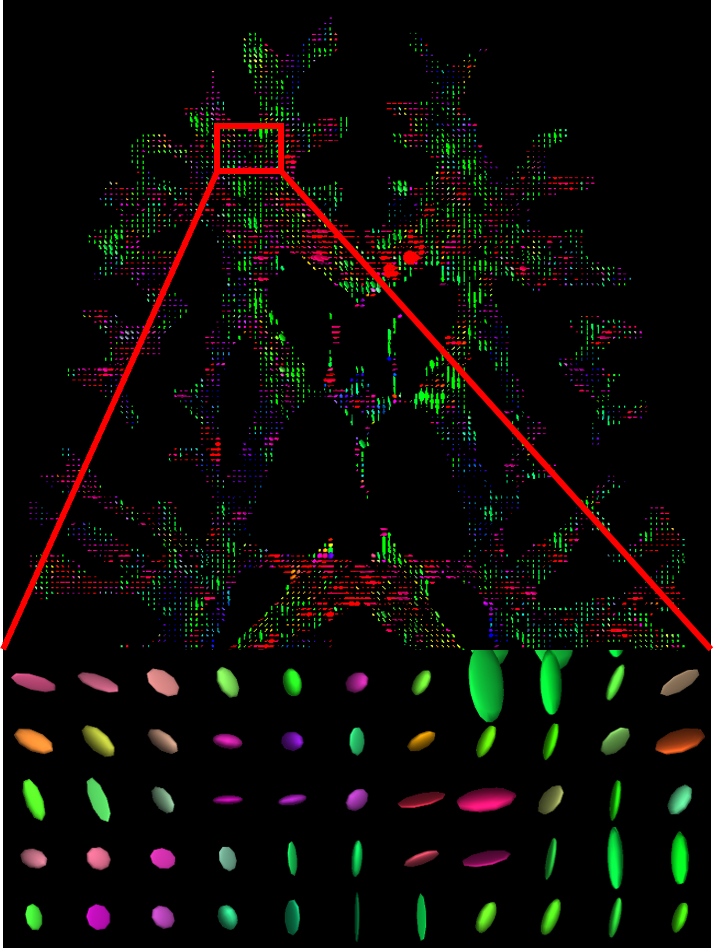}}
 \vspace{0.1cm}
\end{minipage}
\hfill
\begin{minipage}[b]{0.18\linewidth}
  \centering
  \centerline{\includegraphics[width=2.4cm,height=3.0cm]{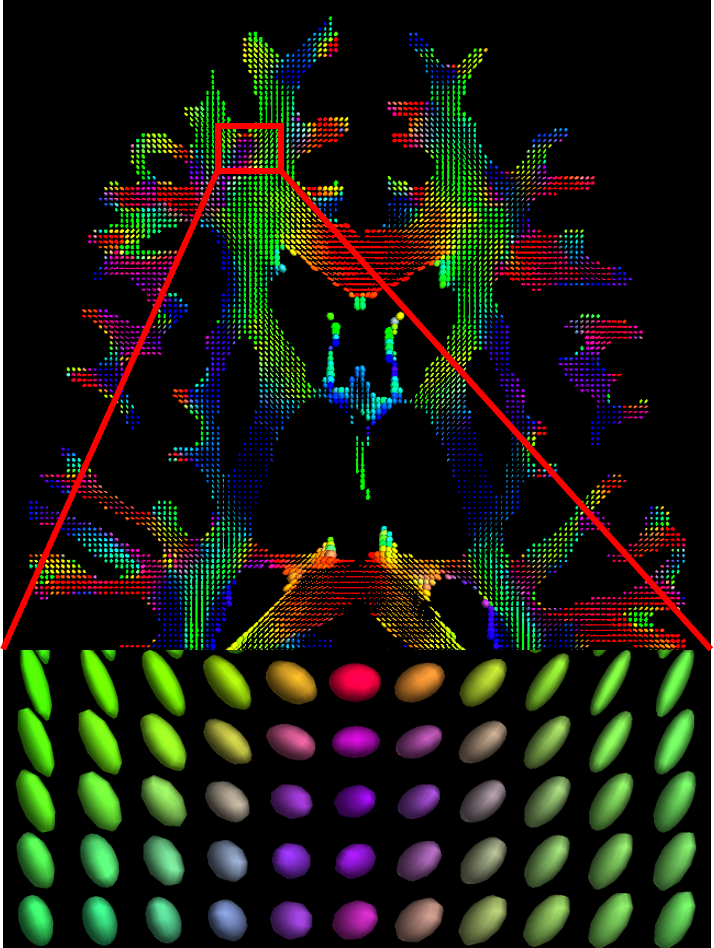}}
 \vspace{0.1cm}
\end{minipage}
\hfill
\begin{minipage}[b]{0.18\linewidth}
  \centering
  \centerline{\includegraphics[width=2.4cm,height=3.0cm]{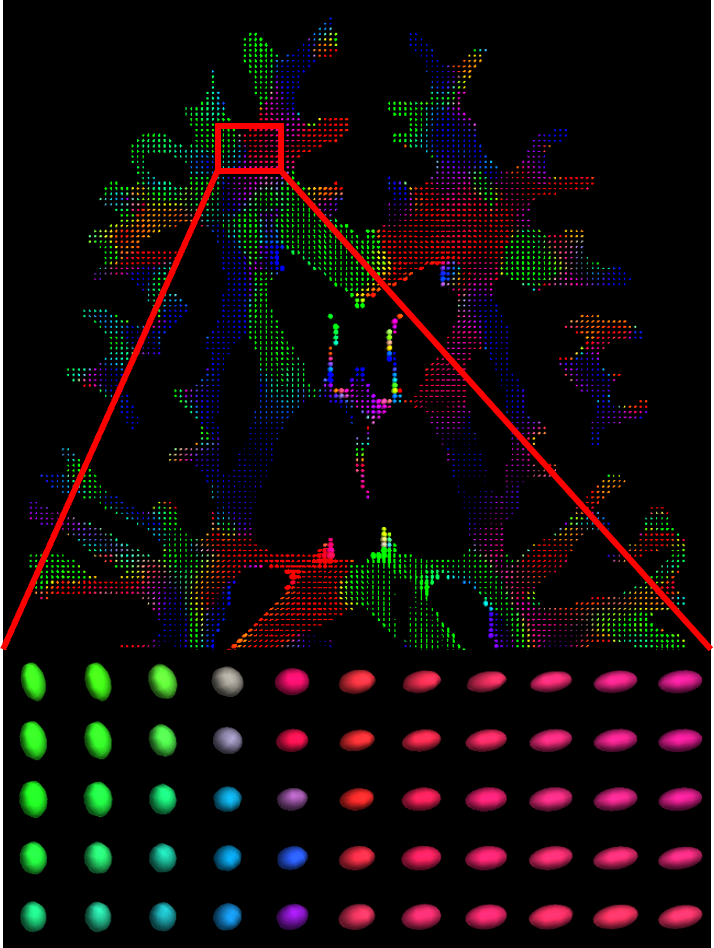}}
 \vspace{0.1cm}
\end{minipage}
\hfill
\begin{minipage}[b]{0.18\linewidth}
  \centering  \centerline{\includegraphics[width=2.4cm,height=3.0cm]{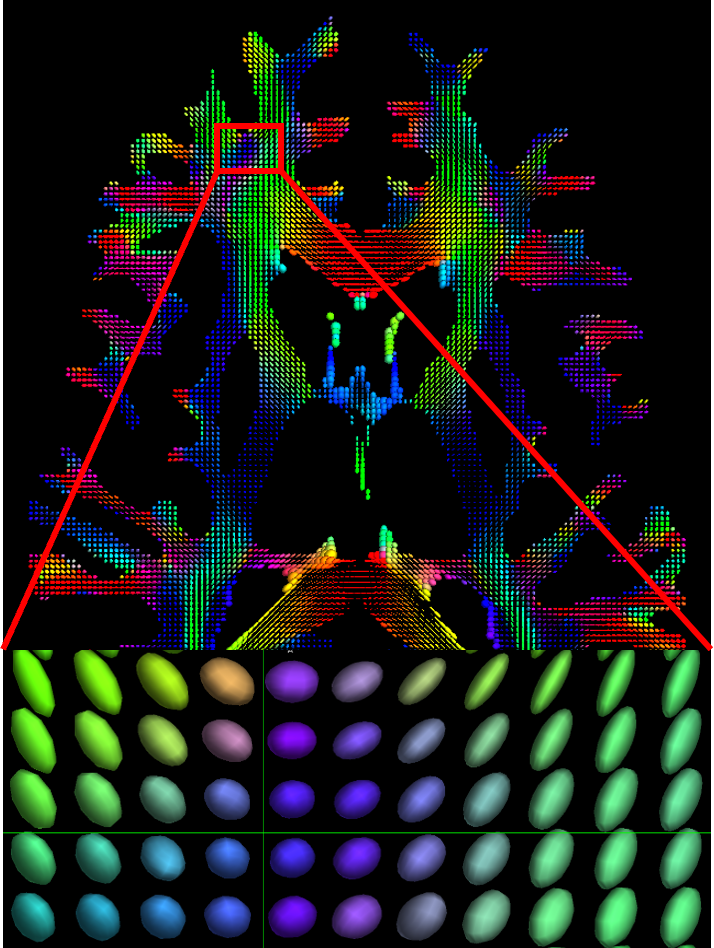}}
 \vspace{0.1cm}
\end{minipage}
\hfill
\begin{minipage}[b]{0.18\linewidth}
  \centering
  \centerline{\includegraphics[width=2.4cm,height=3.0cm]{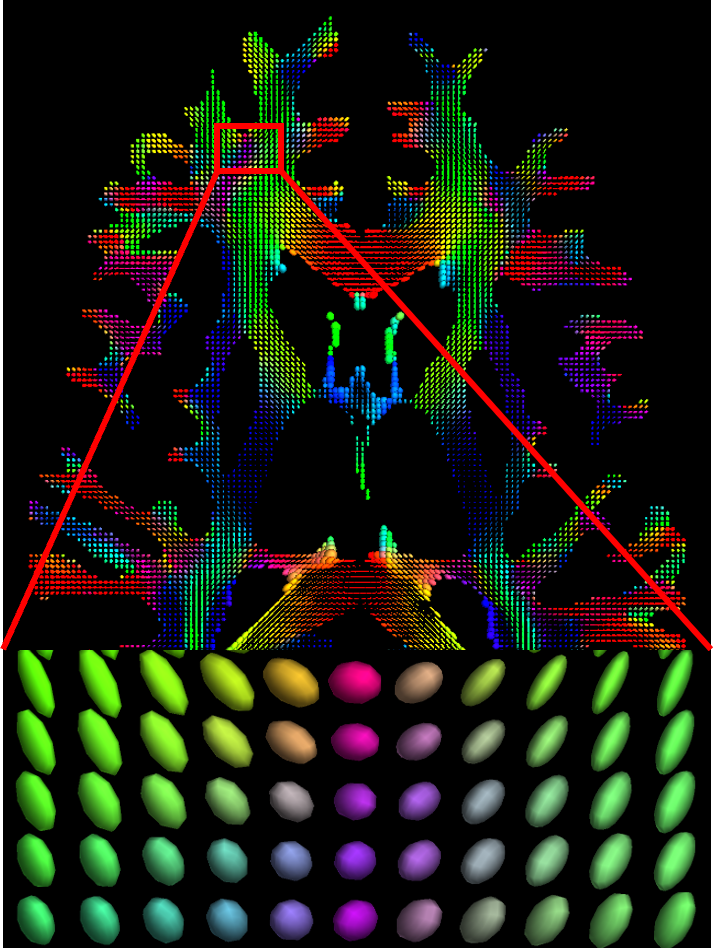}}
 \vspace{0.1cm}
\end{minipage}
\begin{minipage}[b]{0.18\linewidth}
  \centering
  \centerline{\includegraphics[width=2.4cm,height=3.0cm]{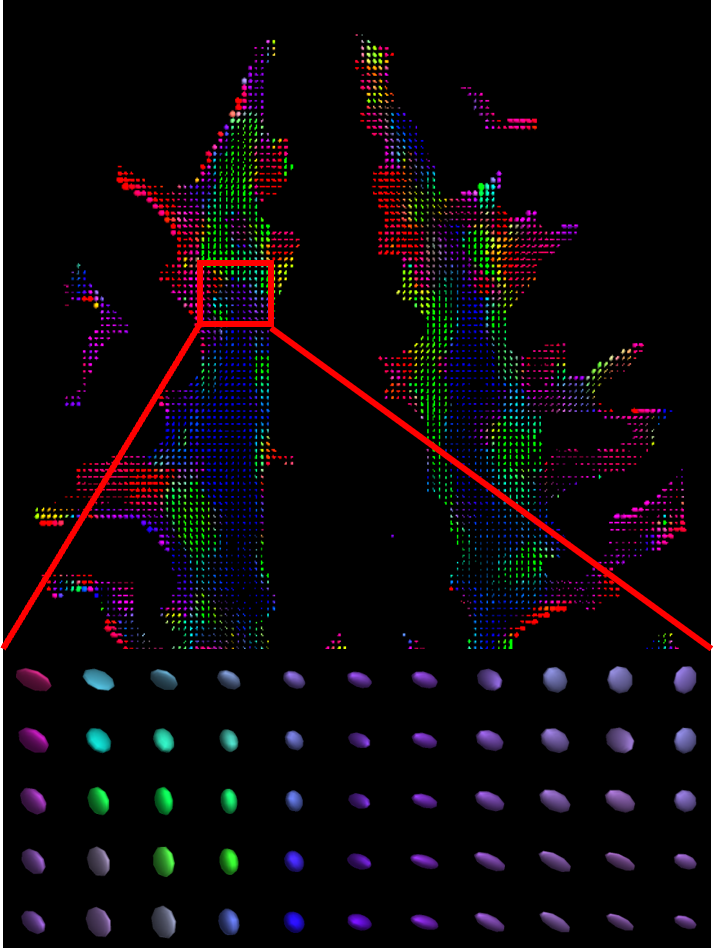}}
  \centerline{(a) 6-dir DTI}\medskip
\end{minipage}
\hfill
\begin{minipage}[b]{0.18\linewidth}
  \centering
  \centerline{\includegraphics[width=2.4cm,height=3.0cm]{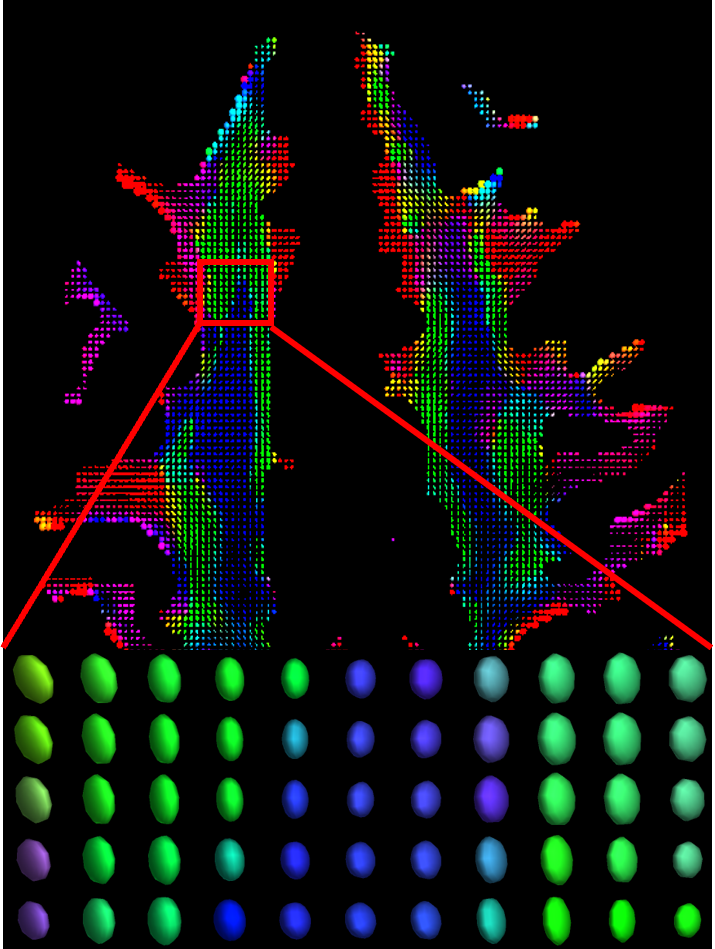}}
  \centerline{(b) GT}\medskip
\end{minipage}
\hfill
\begin{minipage}[b]{0.18\linewidth}
  \centering\centerline{\includegraphics[width=2.4cm,height=3.0cm]{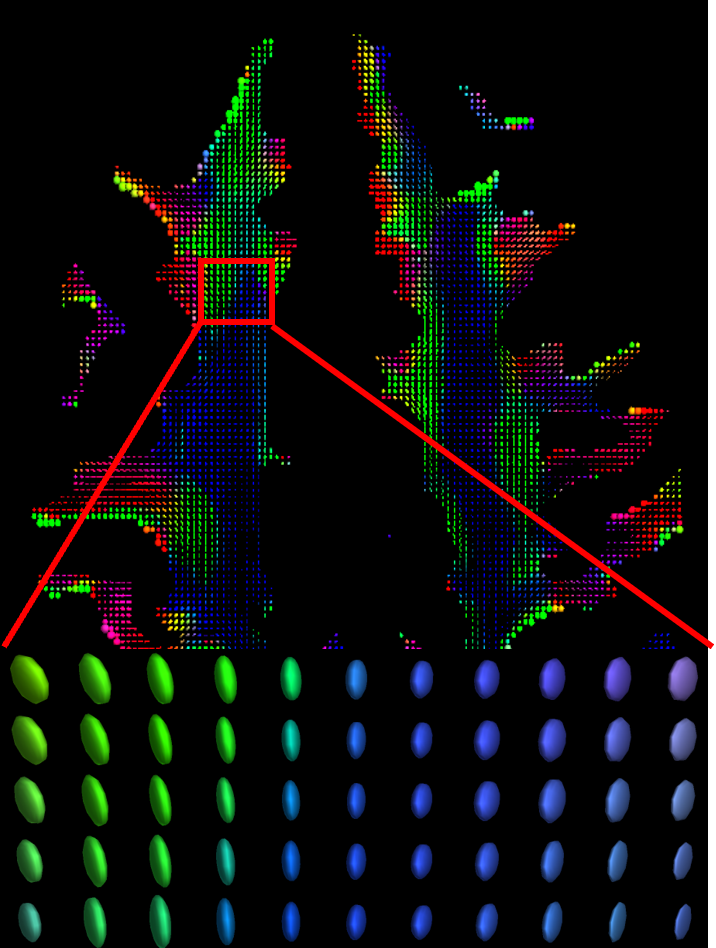}}
  \centerline{(c) DeepDTI}\medskip
\end{minipage}
\hfill
\begin{minipage}[b]{0.18\linewidth}
  \centering
  \centerline{\includegraphics[width=2.4cm,height=3.0cm]{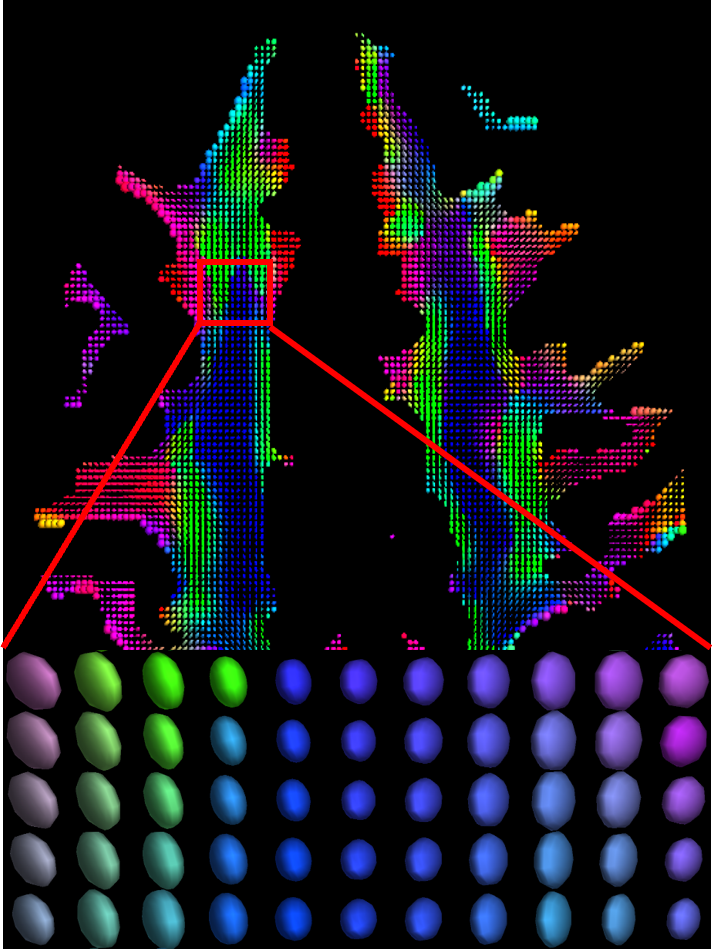}}
  \centerline{(d) HADTI-Net}\medskip
\end{minipage}
\hfill
\begin{minipage}[b]{0.18\linewidth}
  \centering
  \centerline{\includegraphics[width=2.4cm,height=3.0cm]{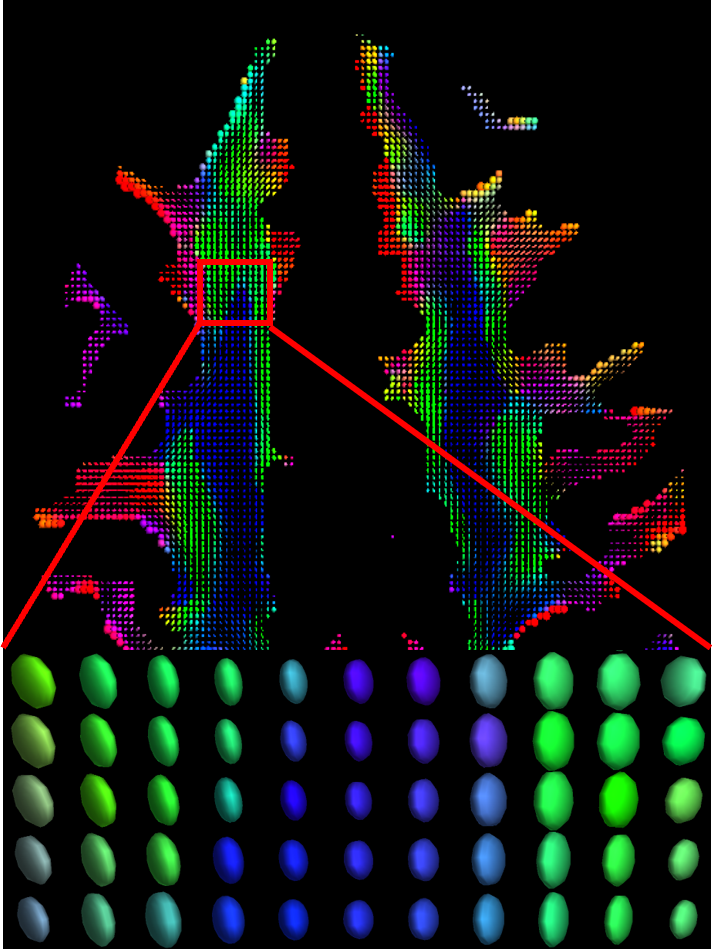}}
  \centerline{\textbf{(e) DirGeo-DTI}}\medskip
\end{minipage}
\caption{Axial visualisations of DTI tensors by different approaches: (a) 6-dir DTI, (b) ground truth, enhanced DTI by (c) DeepDTI, (d) HADTI-Net, and the proposed (e) DirGeo-DTI using the same set of inputs ($b_0$ + 6dir DWIs) for a testing subject in the HCP and PPMI dataset. The first and second rows represent results from the HCP and PPMI, respectively.}\label{fig:tensor_vis}
\end{figure}

From the results, it can be observed that the proposed method achieves the best performance in all the metrics. To further demonstrate the superior performance of the proposed method across all metrics, particularly from the perspective of structural information at both local and global levels, we have calculated and summarised the Structural Similarity Index (SSIM) in Table~\ref{SSIM}. Furthermore, examples of qualitative results for a testing subject from the HCP and PPMI datasets are shown in Figure~\ref{fig:tensor_vis} as two rows, where the first row represents results from the HCP dataset, while the second row shows results from the PPMI dataset.

\subsection{WM Tracts Analysis}
\begin{figure*}[!hbt]
\centering
\includegraphics[width=\linewidth]{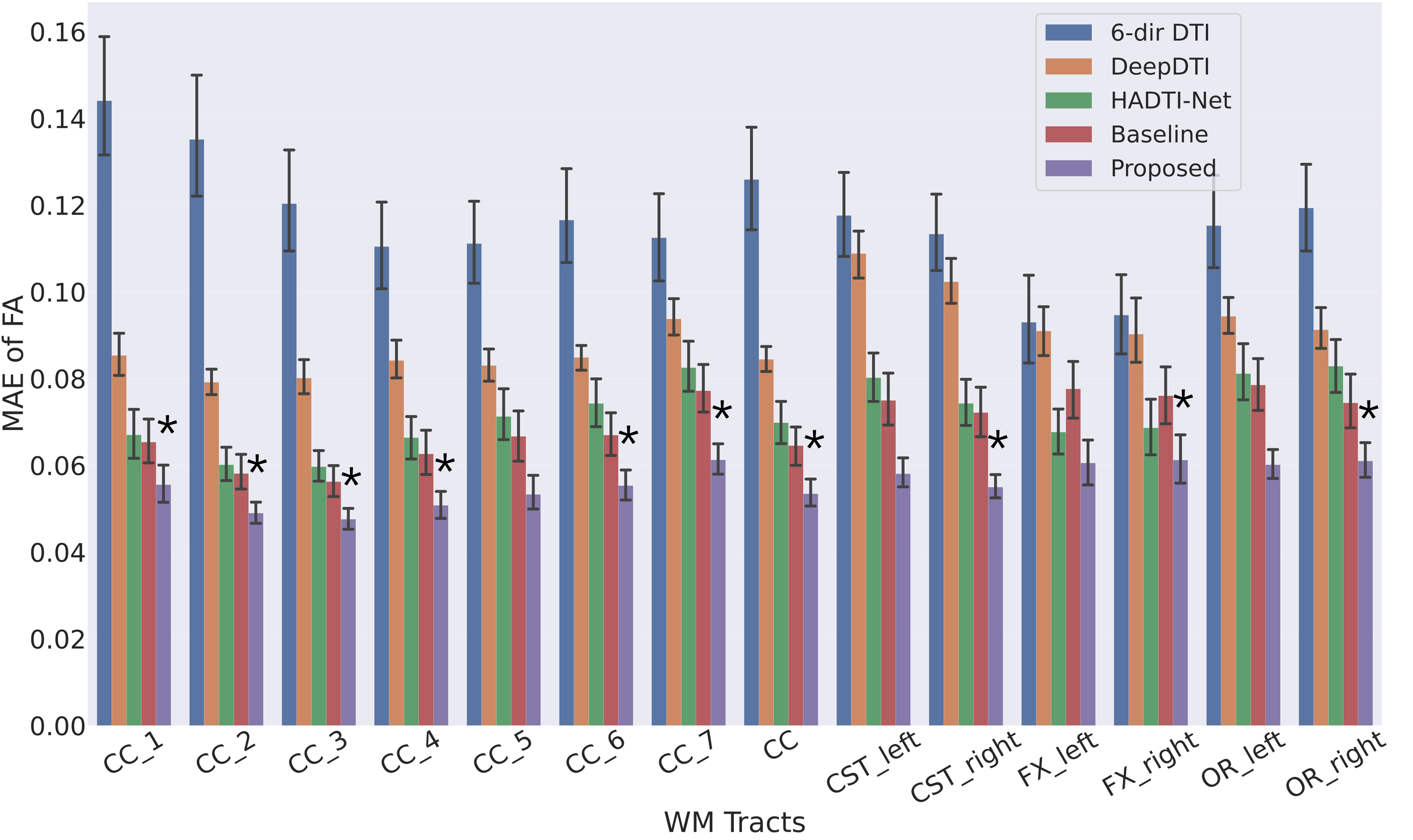}
  \caption{Mean absolute FA differences for 6-dir DTI and enhanced DTI by different methods in specific WM tracts of PPMI testing subjects.  * denotes no statistically significant difference from the ground truth with $p$-value $>$ 0.05.}
  \label{fig:tract_ppmi}
\end{figure*}
\begin{figure*}[!bth]
  \centering
\includegraphics[width=\linewidth]{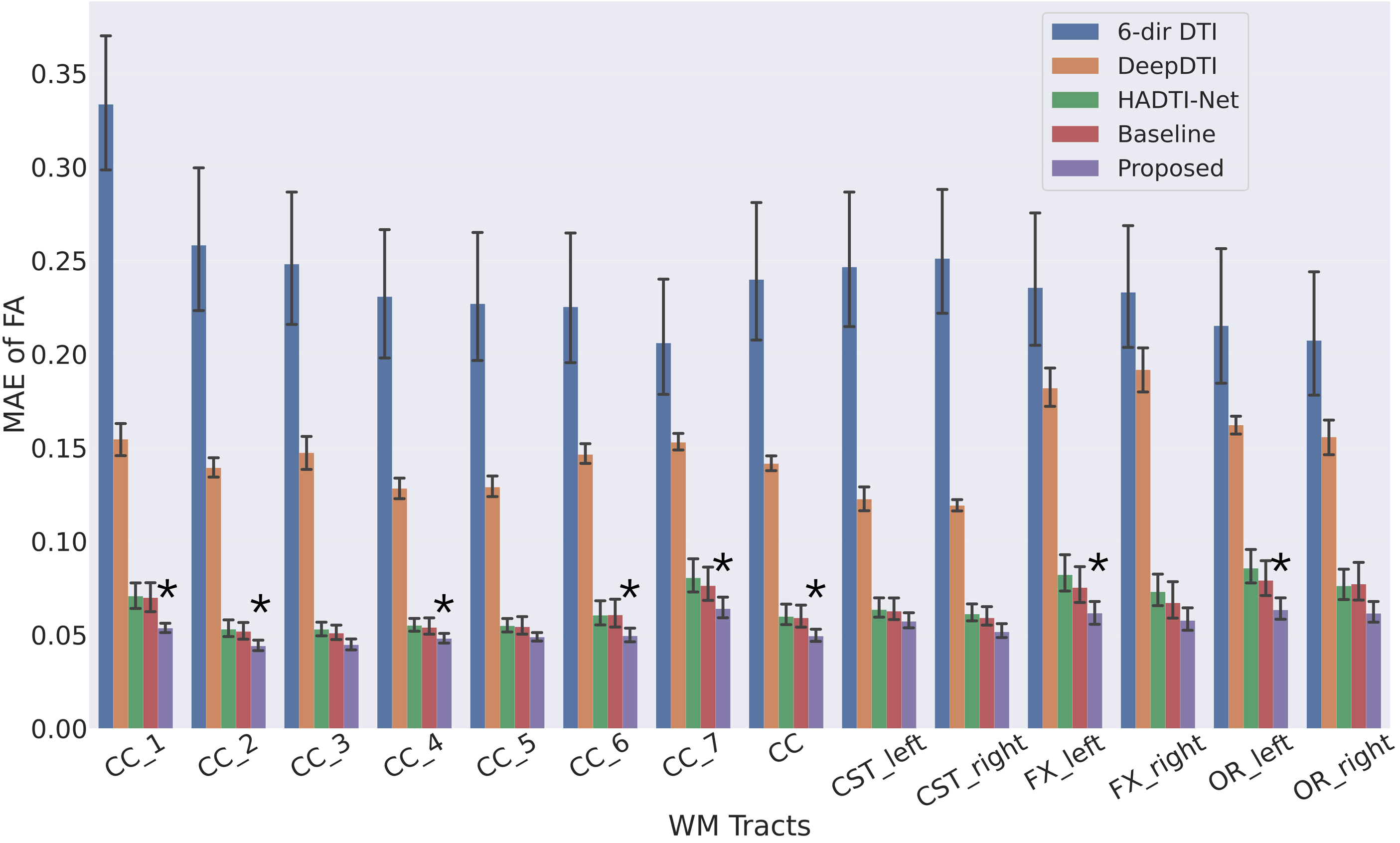}
  \caption{Mean absolute FA differences for 6-dir DTI and enhanced DTI by different methods in specific WM tracts of HCP testing subjects. * denotes no statistically significant difference from the ground truth with $p$-value $>$ 0.05.}
  \label{fig:tract_hcp}
\end{figure*}
WM tracts are crucial in clinical research on neurological diseases due to their role in connecting different regions of the brain, facilitating efficient communication and coordination of neural activities. Understanding the integrity and functionality of white matter tracts is vital for diagnosing and managing neurological conditions. To further explore the robustness and potential clinical significance of the proposed DirGeo-DTI, several WM tracts of interest were selected for further evaluation based on the interests from clinical research in Alzheimer’s disease (AD)~\cite{yasmin2008diffusionUF}, Huntington's disease (HD)~\cite{rosas2006diffusion}, Multiple Sclerosis (MS)~\cite{pujol1997lesionsAF,roosendaal2009regional}, and Parkinson’s disease (PD)~\cite{li2018analysisAF}. The segmentations of WM tracts were performed using TractSeg~\cite{wasserthal2018tractseg}. A repetitive experiment was performed by inferencing the compared methods using 10 different subsets of 6-direction DWIs extracted from the original ground truth acquisition. This experiment compares the $\FA$ differences in the enhanced DTIs predicted by different methods to the $\FA$ differences in WM tracts reported in previous studies. The corresponding results for the testing subjects in PPMI and HCP are shown in Figure~\ref{fig:tract_ppmi} and Figure~\ref{fig:tract_hcp}, respectively. The selected white matter (WM) tract bundles are defined according to~\cite{wasserthal2018tractseg} and include the corpus callosum (CC) with its subregions (rostrum, genu, rostral body, anterior midbody, posterior midbody, isthmus, and splenium), as well as the cingulum (CG), corticospinal tract (CST), fornix (FX), and optic radiation (OR).

The hypothesis is that if the MAE of FA values between the enhanced DTI and the ground truth exceeds clinically observed differences, the likelihood of detecting group-wise differences within the same sample size may decrease significantly. For instance, $\FA$ differences of 0.07 were observed in both the left and right optic radiation tracts in patients with MS when compared to healthy controls~\cite{li2018analysisAF,pujol1997lesionsAF}. The DTIs generated by 6-dir DTI, DeepDTI, and HADTI-Net exhibit FA differences that are either larger than or close to the observed group-wise differences. In contrast, the proposed method effectively reduces measurement errors to a level below the gap between patients and controls, suggesting the potential to still detect these group-wise differences. Although the evaluation results on the enhanced DTI only show no statistical difference of $\FA$ in only 34 out of 72 WM tracts for PPMI (26 out of 72 for HCP) compared to the ground truth, the proposed DirGeo-DTI demonstrates clear mitigation of $\FA$ differences compared to other methods.

\section{Conclusion and Future Work}
We have proposed DirGeo-DTI for enhancing the angular resolution of DTI from DWI data with a minimal number of gradient directions. By leveraging directional information and geometric learning, DirGeo-DTI effectively improves DTI predictions, as demonstrated by the experimental results on both the HPC and PPMI datasets. Our findings suggest that DirGeo-DTI has the potential to be a valuable post-processing tool in clinical research, particularly in studies where only a limited number of diffusion directions are available, thereby enhancing the utility of DWI scans and the reliability of scalar metrics derived from DTI.

While results are promising, further work is needed to fully evaluate DirGeo-DTI in real-world clinical settings. Specifically, we plan to assess the method on actual clinical data with fewer gradient directions and potential artifacts to better understand its robustness and applicability in typical clinical environments. This future direction will provide a broader understanding of the method’s performance across diverse scenarios, ultimately enhancing its utility in clinical practice.

\subsubsection{Acknowledgments}
The authors acknowledge the support of The BISA Flagship Research Program (2019) and The University of Sydney Office of Global and Research Engagement Catalyst Grants (2024).

\bibliographystyle{splncs04}
\bibliography{7915.bib}
\end{document}